\documentclass[11pt,a4paper]{article}
\usepackage{arabtex}
\usepackage{utf8}
\setcode{utf8}
\usepackage[hyperref]{acl2020}
\usepackage{times}
\usepackage{latexsym}
\usepackage{graphicx}
\usepackage{multirow}
\usepackage{booktabs}
\usepackage{subfigure}

\usepackage{microtype}

\usepackage{url}
\usepackage{amssymb}% http://ctan.org/pkg/amssymb
\usepackage{pifont}% http://ctan.org/pkg/pifont
\newcommand{\cmark}{\ding{51}}%
\newcommand{\xmark}{\ding{55}}%

\aclfinalcopy % Uncomment this line for the final submission
\def\aclpaperid{2112} %  Enter the acl Paper ID here

\title{A Multitask Learning Approach for Diacritic Restoration}
\author{Sawsan Alqahtani \textsuperscript{1,2} \and Ajay Mishra\textsuperscript{1} \and Mona  Diab\textsuperscript{2}\thanks{\noindent\textsuperscript{*}The work was conducted while the author was with AWS, Amazon AI.} \\
\textsuperscript{1}AWS, Amazon AI \\
\textsuperscript{2}The George Washington University \\
\texttt{sawsa@amazon.com}, \texttt{misaja@amazon.com}, \texttt{mtdiab@gwu.edu}}
  
\date{}

\begin{document}
\maketitle

\begin{abstract}
    In many languages like Arabic, diacritics are used to specify pronunciations as well as meanings. Such diacritics are often omitted in written text, increasing the number of possible pronunciations and meanings for a word. This results in a more ambiguous text making computational  processing on such text more difficult. Diacritic restoration is the task of restoring missing diacritics in the written text. Most state-of-the-art diacritic restoration models are built on character level information which helps generalize the model to unseen data, but presumably lose  useful information at the word level.  Thus, to compensate for this loss, we investigate the use of multi-task learning to jointly optimize diacritic restoration with related NLP problems namely word segmentation, part-of-speech tagging, and syntactic  diacritization. We  use  Arabic  as  a  case study since it has sufficient data  resources  for  tasks  that  we  consider  in our joint modeling. Our joint models significantly  outperform the baselines and are comparable to the state-of-the-art models that are more complex relying on morphological analyzers and/or a lot more data (e.g. dialectal data).  

\end{abstract} 

%To the best of our knowledge, this study is the first that investigates the benefits of automatically learning related tasks to boost the performance of diacritic restoration.  

\section{Introduction}

In contrast to English, some vowels in languages such as Arabic and Hebrew are not part of the alphabet and diacritics are used for vowel specification.\footnote{Diacritics are marks that are added above, below, or in-between the letters to compose a new letter or characterize the letter with a different sound \citep{wells2000orthographic}.} In addition to pertaining vowels, diacritics can also represent other features such as case marking and phonological gemination in Arabic.
%Other languages that include diacritics such as Yoruba and Vietnamese sometimes omit diacritics in writing for various reasons such as difficulty in typing diacritics on keyboards or digitizing electronic text \citep{scannell2011statistical}. 
Not including diacritics in the written text in such languages increases the number of possible meanings as well as pronunciations. Humans rely on the surrounding context and their previous knowledge  to infer the meanings and/or pronunciations of words. However, computational models, on the other hand, are inherently limited to deal with missing diacritics which pose a challenge for such models due to increased ambiguity.

%The importance of specifying vowels or diacritics is apparent in speech related applications such as speech recognition and text-to-speech \cite{vergyri2004automatic,ungurean2008automatic}. In theory, other semantic or syntactic related NLP applications, such as machine translation and part-of-speech tagging, also benefit from accurate assignment of diacritics. However, it is not clear yet whether the impact of missing diacritics in such applications is significant \cite{alqahtani2016investigating,alqahtani2019homograph}.

Diacritic restoration (or diacritization) is the process of restoring these missing diacritics for every character in the written texts. It can specify pronunciation and can be viewed as a relaxed variant of word sense disambiguation. For example, the Arabic word \<علم>
\textit{Elm}\footnote{We use Buckwalter Transliteration encoding  http://www.qamus.org/transliteration.htm.} can mean ``flag" or ``knowledge", but the meaning as well as pronunciation is specified when the word is diacritized (
\<عَلَمُ>
\textit{E\textbf{a}l\textbf{a}m\textbf{u}}  means ``flag"  while 
\<عِلْمْ> \textit{E\textbf{i}l\textbf{o}m\textbf{o}} means ``knowledge"). %Equivalently, the Yoruba word \textit{mu} when unmarked means ``drink", but when diacritized as \textit{m$\grave{u}$} or \textit{m$\acute{u}$} mean ``sink" or ``sharp", respectively.
As an illustrative example in English, if we omit the vowels in the word \textit{pn}, the word can be read as \textit{pan}, \textit{pin}, \textit{pun},  and \textit{pen}, each of these variants have different pronunciations and meanings if it composes a valid word in the language.

The state-of-the-art diacritic restoration models reached a decent performance  over the years using recurrent or convolutional neural networks in terms of accuracy \cite{zalmout2017don,sawsantcn,orife2018attentive} and/or efficiency \cite{sawsantcn,orife2018attentive}; yet, there is still room for further improvements. Most of these models are built on character level information which help generalize the model to unseen data, but presumably lose some useful information at the word level. Since word level resources are insufficient to be relied upon for training diacritic restoration models, we integrate additional linguistic information that considers word morphology as well as word relationships within a sentence to partially compensate for this loss.

In this paper, we improve the performance of diacritic restoration  by building a multitask learning model (i.e. joint modeling). Multitask learning refers to models that learn more than one task at the same time, and has recently been shown to provide good solutions for a number of NLP tasks \cite{hashimoto2016joint,kendall2018multi}. %Reviewer comment: it alleviates reliance on external resources, however, if these resources already exist, why should they not be used. To over-simplify a little, the authors have simply taken one step out of the preprocessing pipeline and lumped it in to the end-to-end system, which is fine, but the contribution seems to be over-advertised in the introduction. 

%The use of a multitask learning approach provides a convenient and better alternative than generating all linguistic features as a preprocessing step for diacritic restoration. In addition, it alleviates the reliance on other computational and/or data resources to generate these features. 
The use of a multitask learning approach provides an end-to-end solution, in contrast to generating the linguistic features for diacritic restoration as a preprocessing step. In addition, it alleviates the reliance on other computational and/or data resources to generate these features. Furthermore, the proposed model is flexible such that a task can be added or removed depending on the data availability. This makes the model adaptable to other languages and dialects.

We consider the following auxiliary tasks to boost the performance of diacritic restoration: word segmentation, part-of-speech (POS) tagging, and syntactic diacritization. We use Arabic as a case study for our approach since it has sufficient data resources for tasks that we consider in our joint modeling.\footnote{Other languages that include diacritics lack such resources; however, the same multitask learning framework can be applied if data resources become available.}

The contributions of this paper are twofold: 
\vspace{-2 mm} 
\begin{enumerate}
    \item We investigate the benefits of automatically learning related tasks to boost the performance of diacritic restoration;
    \vspace{-2 mm} 
    \item In doing so, we devise a state-of-the-art model for Arabic diacritic restoration as well as a framework for improving diacritic restoration in other languages that include diacritics.  
\end{enumerate}

\section{Diacritization and Auxiliary Tasks}
\label{aux_tasks}

We formulate the problem of \textit{(full)} diacritic restoration  (\textit{DIAC}) as follows: given a sequence of characters, we identify the diacritic corresponding to each character in that sequence from the following set of diacritics \{a, u, i, o, K, F, N, $\sim$, $\sim$a, $\sim$u, $\sim$i, $\sim$F, $\sim$K, and $\sim$N\}. We additionally consider three auxiliary tasks: syntactic diacritization, part-of-speech tagging, and word segmentation. Two of which operate at the word level (syntactic diacritization and POS tagging) and the remaining tasks (diacritic restoration and word segmentation) operate at the character level. This helps diacritic restoration utilize information from both character and word level information, bridging the gap between the two levels. 

\paragraph{Syntactic Diacritization \textit{(SYN)}:} This refers to the task of retrieving diacritics related to the syntactic positions for each word in the sentence, which is a sub-task of full diacritic restoration. Arabic is a templatic language where words comprise roots and patterns in which patterns are typically reflective of diacritic distributions. Verb patterns are more or less predictable however nouns tend to be more complex. Arabic diacritics can be divided into lexical and inflectional (or syntactic) diacritics. Lexical diacritics change the meanings of words as well as their pronunciations and their distribution is bound by patterns/templates. In contrast, inflectional diacritics are related to the syntactic positions of words in the sentence and are added to the last letter of the main morphemes of words (word finally), changing  their pronunciations.\footnote{Diacritics that are added due to passivization are also syntactic in nature but are not considered in our syntactic diacritization task. That said, they are still considered in the full diacritic restoration model.} Inflectional diacritics are also affected by word's root (e.g. weak roots) and semantic or morphological properties (e.g. with the same grammatical case, masculine and feminine plurals take different diacritics).

Thus, the same word can be assigned a different syntactic diacritic reflecting syntactic case, i.e. depending on its relations to the remaining words in the sentence (e.g. subject or object). For example, the diacritized variants 
\<عَلَمَ>
Ealam\textbf{a} and 
\<عَلَمُ>
Ealam\textbf{u} which both mean ``flag" have the corresponding syntactic diacritics: \textbf{a} and \textbf{u}, respectively. That being said, the main trigger for accurate syntactic prediction is the relationships between words, capturing semantic and most importantly, syntactic information.

Because Arabic has a unique set of  diacritics, this study formulates syntactic diacritization in the following way: each word  in the input is tagged with a single diacritic  representing its syntactic position in the sentence.\footnote{Combinations of diacritics is possible but we combine valid possibilities together as one single unit in our model. For example, the diacritics \textbf{$\sim$} and \textbf{a} are combined to form an additional diacritic \textbf{$\sim$a}.} The set of diacritics in syntactic diacritization is the same as the set of diacritics for full diacritic restoration. Other languages that include diacritics can include syntactic related diacritics but in a different manner and complexity compared  to Arabic.

\paragraph{Word segmentation \textit{(SEG)}:} This refers to the process of separating affixes  from the main unit of the word. Word segmentation is commonly used as a preprocessing step for different NLP applications and its usefulness is apparent in morphologically rich languages.  For example, the undiacritized word \textit{whm} \<وهم> might be diacritized as  \textit{waham$\sim$a} \<وَهَمَّ> ``and concerned", \textit{waham} \<وَهَم> ``illusion", where the first diacritized word consists of two segments ``wa ham$\sim$a" \<وَ  هَمَّ> while the second is composed of one word. Word segmentation can be formulated in the following way: each character in the input is tagged following IOB tagging scheme (\textit{B}: beginning of a segment; \textit{I}: inside a segment; \textit{O}: out of the segment) \cite{diab2004automatic}. %%SA cite my work on tokenization using IOB since my work established this approach Diab et al 2004. DONE 

\paragraph{Part-Of-Speech Tagging \textit{(POS)}:}  This refers to the task of determining the syntactic role of a word (i.e. part of speech) within a sentence. POS tags are highly correlated with diacritics (both syntactic and lexical): knowing one helps determine or reduce the possible choices of the other. For instance, the word \<كتب> \textit{ktb} in the sentence \textit{ktb [someone]} means ``books" if we know it to be a noun whereas the word would be either \<كَتَب> \textit{katab} ``someone wrote" or \<كَتَّب> \textit{kat$\sim$ab} ``made someone write" if it is known to be a verb. 

POS tagging can be formulated in  the following way: each word in the input is assigned a POS tag from the Universal Dependencies tagset \cite{taji2017universal}.\footnote{Refer to https://universaldependencies.org/. This tagset is chosen because it includes essential POS tags in the language, and it is unified across different languages which makes it suitable to investigate more languages in the future. }

\section{Approach}

We built a diacritic restoration joint model and studied   the extent to which sharing information is plausible to improve diacritic restoration performance. Our joint model is motivated by the recent success of the hierarchical modeling proposed in \cite{hashimoto2016joint} such that information learned from an auxiliary task is passed as input to the diacritic restoration related layers.\footnote{We also experimented with learning tasks sharing some levels and then diverging to specific layers for each tasks. However, this did not improve the performance compared to the diacritic restoration model when we don't consider any additional task.} 

\subsection{Input Representation}
\label{input_representation}
\begin{figure}[h]
\centering
  \includegraphics[width=5cm,height=5cm]{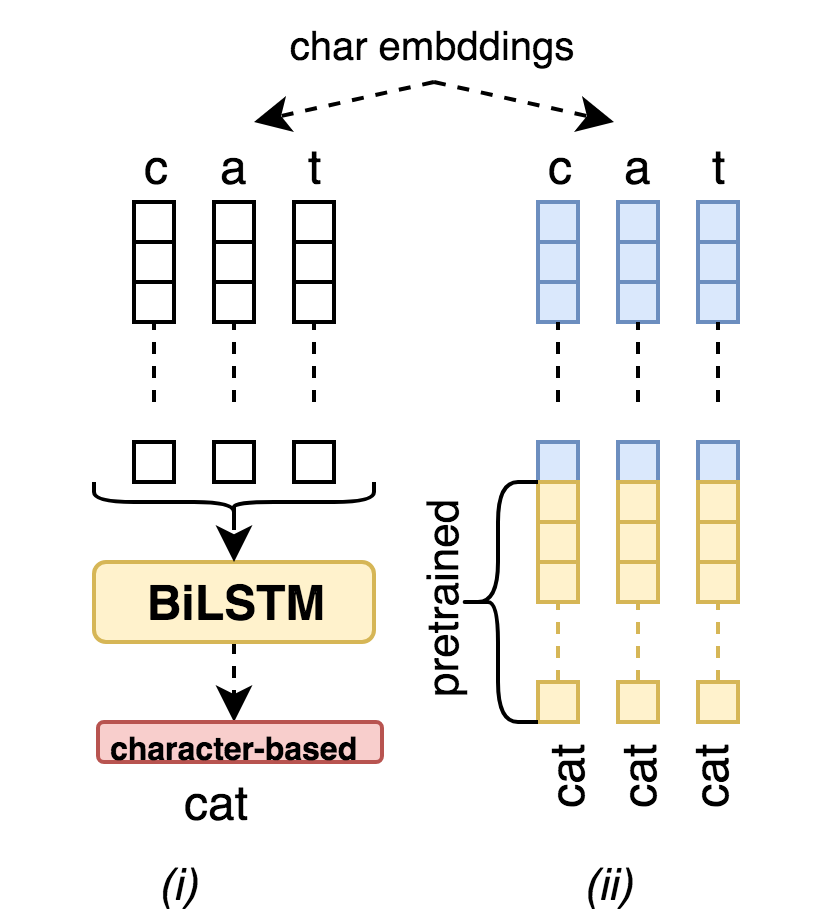}
  \caption{\label{charactertoword}An example of embedding vectors for the word \textit{cat} and its individual characters: c,a, and t. (i) A character-based representation for the word \textit{cat} from its individual characters; (ii) A concatenation for the word embedding with each of its individual characters.}
  \vspace{-4 mm} 

\end{figure}

Since our joint model may involve both character and word level based tasks, we began our investigation by asking the following question: \textit{how to integrate information between these two levels?} Starting from the randomly initialized character embeddings as well as a pretrained set of embeddings for words, we follow two approaches (Figure \ref{charactertoword} visually illustrates the two approaches with an example).

\begin{figure*}[h]
\centering
  \includegraphics[width=12cm,height=8cm]{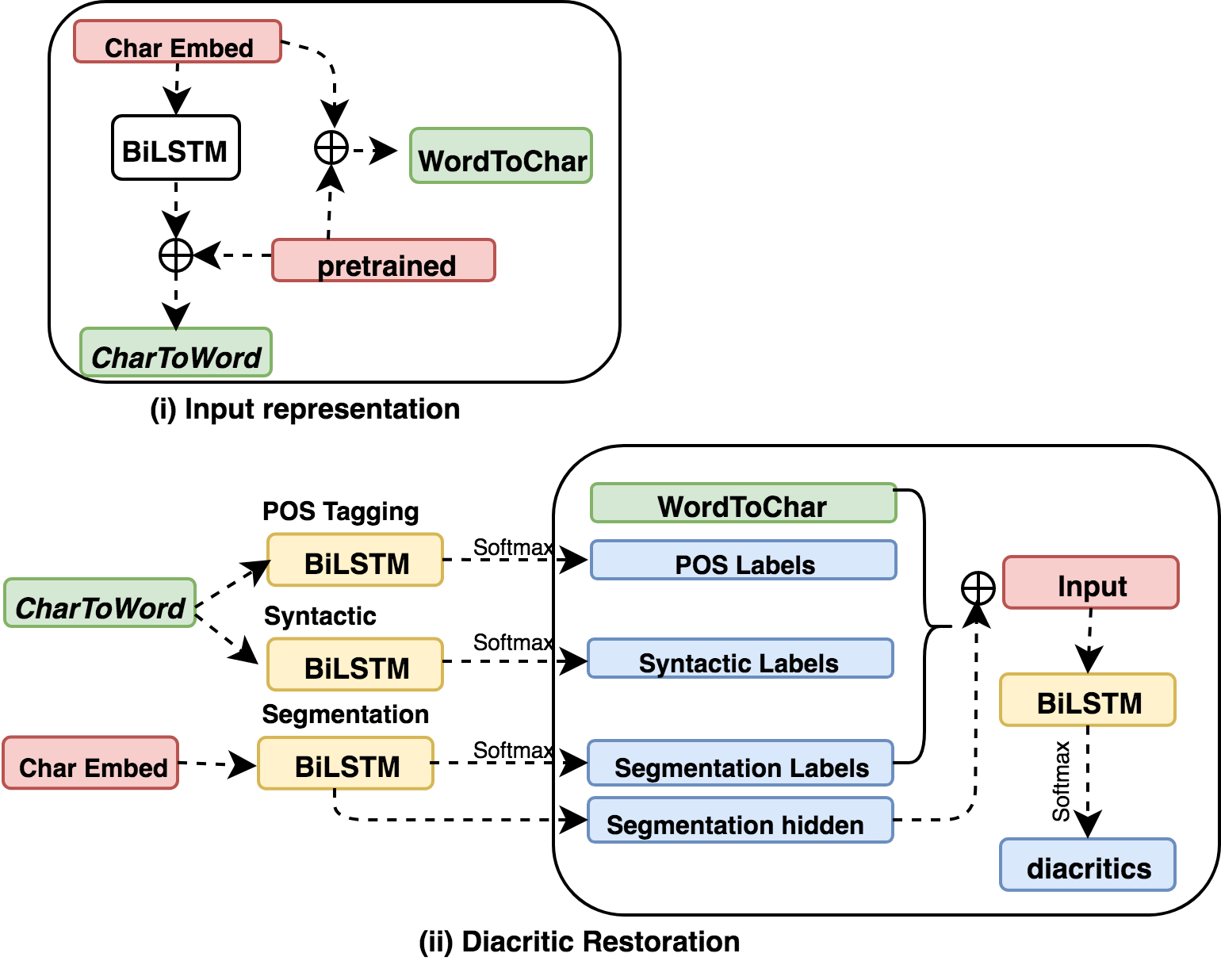}
  \caption{The diacritic restoration joint model. All \textit{Char Embed} entities refer to the same randomly initialized character embedding learned during the training process. \textit{Pretrained} embeddings refer to fixed word embeddings obtained from fastText \cite{bojanowski2017enriching}. \textit{(i)} shows the input representation for CharToWord and WordToChar embedding which is the same as in Figure \ref{charactertoword}. \textit{(ii)} represents the diacritic restoration joint model; output labels from each task are concatenated with WordToChar embedding and optionally with segmentation hidden.}
  \label{architecture}
    \vspace{-4 mm} 

\end{figure*}

\paragraph{\textit{(1) Character Based Representation}:} We pass information learned by character level tasks into word level tasks by composing a word embedding from the word's characters. We first concatenate the individual embeddings of characters in that word, and then apply a Bidirectional Long Short Term Memory (BiLSTM) layer to generate denser vectors.\footnote{We also evaluated the use of a feedforward layer and  unidirectional Long Short Term Memory (LSTM) but a BiLSTM layer yielded better results.} This helps representing morphology and word composition into the model. 
    
%%name of the autor then the year at the end 

\paragraph{\textit{(2) Word-To-Character Representation}:} To pass information learned by word level tasks into character level tasks, we concatenate each word with each of its composed characters during each pass, similar to what is described in \citet{watson2018utilizing}'s study. This helps distinguishing the individual characters based on the surrounding context, implicitly capturing additional semantic and syntactic information.  

\subsection{The Joint Model}

For all architectures, the main component is BiLSTM \cite{hochreiter1997long,schuster1997bidirectional}, which preserves the temporal order of the sequence and has been shown to provide the state-of-the-art performance in terms of accuracy \cite{zalmout2017don,sawsantcn}. After representing characters through random initialization and representing words using pretrained embeddings obtained from fastText \cite{bojanowski2017enriching}, the learning process for each batch runs as  follows: 
\begin{enumerate}
    \item We extract the two additional input representation described in Section
\ref{input_representation}; 
\vspace{-2 mm} 

    \item We apply BiLSTM for each of the different tasks separately to obtain their corresponding outputs;
    \vspace{-2 mm} 

    \item We pass all outputs from all tasks as well as WordToChar embedding vectors as input to the diacritic restoration model and obtain our diacritic outputs. 
\end{enumerate}

Figure \ref{architecture} illustrates the diacritic restoration joint model. As can be seen, SYN as well as POS tagging are trained on top of CharToWord representation which is basically the concatenation of the pretrained embedding for each word with the character-based representations described in Figure \ref{charactertoword}. SEG is also trained separately on top of the character embeddings. We pass the outputs of all these tasks along with WordToChar representation to train the BiLSTM diacritic restoration model.  Omitting a task is rather easy, we just remove the related components for that task to yield the appropriate model. We optionally pass the last hidden layer for SEG along with the remaining input to the diacritic restoration model.\footnote{Passing the last hidden layer for POS tagging and/or SYN did not improve the performance; the pretrained embeddings are sufficient to capture important linguistic signals.}

\section{Experimental Setups}

\paragraph{Dataset:}
We use the Arabic Treebank (ATB) dataset: parts 1, 2, and 3 and follow the same data division as \citet{diab2013ldc}. Table \ref{dataset} illustrates the data statistics. For word based tasks, we segment each sentence into space tokenized words. For character based tasks, we, in addition, add the special boundary ``$\<$w$>$" between these words, and then each word is further segmented into its characters, similar to that in \cite{sawsantcn}. We pass each word through the model along with a specific number of previous and future words (+/- 10 words). %%SA do you mean here units or across the board it is words? --- words across the board -- DONE

\vspace{-2 mm} 

\begin{table}[h]
\small
\centering
\begin{tabular}{ cccc }		
   \textbf{Train} & \textbf{Test} & \textbf{Dev} & \textbf{OOV}\\
  \hline
   502,938 &63,168 & 63,126 & 7.3\%\\
   \hline 
\end{tabular}
  \caption{\label{dataset} Number of words and out of vocabulary (OOV) rate for Arabic. OOV rate indicates the percentage of undiacritized words in the test set that have not been observed during training.}
  \vspace{-4mm}%Put here to reduce too much white space after your table 
\end{table}

  \vspace{-2mm}

\paragraph{Parameter Settings:}
For all tasks, we use 250 hidden units in each direction (500 units in both directions combined) and 300 as embedding size. We use 3 hidden layers for tasks except in SEG in which we use only one layer. We use Adam for learning optimization with a learning rate of 0.001. We use 20 for epoch size, 16 for batch size, 0.3 for hidden dropout, and 0.5 for embedding dropout. We initialize the embedding with a uniform distribution [-0.1,0.1] and the hidden layers with normal distribution.  The loss scores for all considered tasks are combined and then normalized by the number of tasks in the model.  

\paragraph{Evaluation metrics:}
We use accuracy for all tasks except diacritic restoration. For diacritic restoration, the two most typically used metrics are Word Error Rate (WER) and Diacritic Error Rate (DER), the percentages of incorrectly diacritized words and characters, respectively. In order to approximate errors in the syntactic diacritics, we use Last Diacritic Error Rate (LER), the percentage of words that have incorrect diacritics in the last positions of words. To evaluate the models' ability to generalize beyond observed data, we compute WER on OOV (out-of-vocabulary) words.\footnote{Words that appear in the training dataset but do not appear in the test dataset.}

\paragraph{Significance testing:}
We ran each experiment three times and reported the mean score.\footnote{Higher number of experiments provide more robust conclusion about the models' performance. We only considered the minimum acceptable number of times to run each experiment due to limited computational resources.} We used the t-test  with $p=0.05$ to evaluate whether the difference between models' performance and the diacritic restoration is significant \cite{dror2018hitchhiker}.

\section{Results and Analysis}
Table \ref{multi:results} shows the performance of  joint diacritic restoration  models when different tasks are considered. When we consider WordToChar as input to the diacritic restoration model, we observe statistically significant improvements for all evaluation metrics. This is justified by the ability of word embeddings to capture syntactic and semantic information at the sentence level. The same character is disambiguated in terms of the surrounding context as well as the word it appears in (e.g. the character \textit{t} in the word \textit{cat} would be represented slightly different than \textit{t} in a related word \textit{cats} or even a different word \textit{table}). 
We consider both character based model as well as WordToChar based model as our baselines (BASE).

\begin{table*}[h]
\small
\centering
\begin{tabular}{ l|cccc  }		
   \textbf{Task} & \textbf{WER} &  \textbf{DER} & \textbf{LER/Lex} & \textbf{OOV WER}   \\
   \midrule[1pt]
    \newcite{zalmout2017don} & 8.21 & - & - & \textbf{20.2} \\
%    BiLSTM\_CRF & 92.40 \\
    \newcite{zalmout2019adversarial} & \textbf{7.50} & - & - & - \\
    \newcite{alqahtani2019investigating} & 7.6 & 2.7 & - & 32.1 \\

    \midrule[1pt]
   BASE (Char) & 8.51 ($\pm{0.01}$) & 2.80  & 5.20/5.54 & 34.56  \\
   BASE (WordToChar) & 8.09 ($\pm{0.05}$) & 2.73 & 5.00/5.30 &	32.10  \\
  \midrule[0.5pt]
   DIAC+SEG & 8.35 ($\pm{0.02}$)& 2.82 & 5.20/5.46 &	33.97   \\
   DIAC+SYN & 7.70* ($\pm{0.02}$) & 2.60 & 4.72/5.08 & \textbf{30.94}  \\
    DIAC+POS & 7.86* ($\pm{0.14}$) & 2.65 & 4.72/5.20 &	32.28 \\
    \midrule[1pt]
    DIAC+SEG+SYN & 7.70* ($\pm{0.05}$) & 2.59 & 4.65/5.03 & 31.33   \\
    DIAC+SEG+POS & 7.73* ($\pm{0.08}$) &  2.62 & 4.73/5.01  & 31.31  \\ %change WER and SD
    DIAC+SYN+POS  & 7.72* ($\pm{0.06}$) & 2.61 & 4.62/5.06  & 31.05  \\ %change WER and SD
    \midrule[1pt]
    ALL &  \textbf{7.51}* ($\pm{0.09}$) & \textbf{2.54} & \textbf{4.54/4.91}  & 31.07  \\
   \midrule[1pt] 
\end{tabular}
  \caption{\label{multi:results} Performance of the joint diacritic restoration model when different related tasks are considered. \textbf{Bold} numbers represent the highest score per column. Almost all scores are higher than the base model \textit{BASE (char)}. * denotes statistically significant improvements compared to the baselines. \textit{Lex} refers to the percentage of words that have incorrect lexical diacritics only, excluding syntactic diacritics.}
           \vspace{-4mm} 
\end{table*}

We use WordToChar representation rather than characters for all remaining models that jointly learn more than one task. For all experiments, we observe improvements compared to both baselines across all evaluation metrics. Furthermore, all models except DIAC+SEG outperform WordToChar diacritic restoration model in terms of WER, showing the benefits of considering output distributions for the other tasks. Despite leveraging tasks focused on syntax (SYN/POS) or morpheme boundaries (SEG), the improvements extend to lexical diacritics as well. Thus, the proposed joint diacritic restoration model is also helpful in settings beyond word final syntactic related diacritics. The best performance is achieved when we consider all auxiliary tasks within the diacritic restoration model. %Even with the availability of only some related tasks, diacritic restoration still benefit from additional linguistic information.  

\paragraph{Impact of Auxiliary Tasks:}
We discuss the impact of adding each investigated task towards the performance of the diacritic restoration model. 
\subparagraph{Word segmentation (DIAC+SEG):}
When morpheme boundaries as well as diacritics are learned jointly, the WER performance is slightly reduced on all and OOV words. This reduction is attributed mostly to lexical diacritics. %Since morphemes are the building blocks for words and diacritic variations are expected to appear in such morphemes, we accordingly expected to gain more from joint learning morpheme boundaries and diacritics. 
As Arabic exhibits a non-concatenative fusional morphology, reducing its complexity to a segmentation task might inherently obscure  morphological processes for each form.

Observing only slight improvement is surprising; we believe that this is due to our experimental setup and does not negate the importance of having morphemes that assign the appropriate diacritics. We speculate that the reason for this is that we do not capture the interaction between morphemes as an entity, losing some level of morphological information.

For instances, the words \textit{\textbf{w}a\textbf{h}a\textbf{m}$\sim$a} versus \textit{\textbf{w}a\textbf{h}u\textbf{m}} for the undiacritized words \textit{\textbf{whm}} (bold letters refer to consonants distinguishing it from diacritics) would benefit from morpheme boundary identifications to tease apart \textit{\textbf{w}a} from \textit{\textbf{h}u\textbf{m}} in the second variant (\textit{\textbf{w}a\textbf{h}u\textbf{m}}), emphasizing that these are two words. But on the other hand, it adds an additional layer of ambiguity for other cases like the morpheme \textit{\textbf{ktb}} in the diacritic variants \textit{\textbf{k}a\textbf{t}a\textbf{b}a}, \textit{\textbf{k}u\textbf{t}u\textbf{b}u}, \textit{saya\underline{\textbf{k}o\textbf{t}u\textbf{b}o}} - note that the underlined segment has the same consonants as the other variants - in which identifying morphemes increased the number of possible diacritic variants without learning the interactions between adjacent morphemes. 

Furthermore, we found inconsistencies in the dataset for morphemes which might cause the drop in performance when we only consider SEG. When we consider all tasks together, these inconsistencies are reduced because of the combined information from different linguistic signals towards improving the performance of the diacritic restoration model.

\subparagraph{Syntactic diacritization (DIAC+SYN):}
By enforcing inflectional diacritics through an additional focused layer within the diacritic restoration model, we observe improvements on WER compared to the baselines. We notice improvements on syntactic related diacritics (LER score), which is expected given the nature of syntactic diacritization in which it learns the underlying syntactic structure to assign the appropriate syntactic diacritics for each word. Improvements also extend to lexical diacritics, and this is because word relationships are captured during learning syntactic diacritics in which BiLSTM modeling for words is integrated.

\subparagraph{POS tagging (DIAC+POS):}
When we jointly train POS tagging with full diacritic restoration, we notice improvements compared to both baselines. Compared to syntactic diacritization, we obtain \textit{similar} findings across all evaluation metrics except for WER on OOV words in which POS tagging drops. Including POS tagging within diacritic restoration also captures important information about the words; the idea of POS tagging is to learn the underlying syntax of the sentence. In comparison to syntactic diacritization, it involves different types of information like passivization which could be essential in learning correct diacritics.

\subparagraph{Ablation Analysis:}
Incorporating all the auxiliary tasks under study within the diacritic restoration model (ALL) provides the best performance across all measures except WER on OOV words in which the best performance was given by DIAC+SYN. We discuss the impact of removing one task at a time from ALL and examine whether its exclusion significantly impacts the performance. Excluding SEG from the process drops the performance of diacritic restoration. This shows that even though SEG did not help greatly when it was combined solely with diacritic restoration, the combinations of SEG and the other word based tasks filled in the gaps that were missing from just identifying morpheme boundaries. Excluding either POS tagging or syntactic diacritization also hurts the performance which shows that these tasks complement each other and, taken together, they improve the performance of diacritic restoration model. 

\paragraph{Input Representation:}

%\subparagraph{WordToChar versus character representation:} Table \ref{WordToCharImpact} shows the performance of diacritic restoration models when we consider  character embeddings (character only) versus WordToChar representation, without the remaining output labels. Word

\if{false}	
\begin{table}[h!]
\small
\centering
\begin{tabular}{lccc }
Task &  WordToChar &  character only & \\
\midrule[1pt]
    DIAC+SYN  &  \textbf{7.99} & 8.60  \\
    DIAC+POS & \textbf{7.93} & 8.53  \\
    
    \midrule[1pt]
\end{tabular}
  \caption{\label{WordToCharImpact} WER performance when we consider WordToChar representation versus character representation without considering output labels from investigated tasks.}
\end{table}
\fi	

\subparagraph{Impact of output labels:} 
Table \ref{joint:labelnolabel} shows the different models when we do not pass the labels of the investigated tasks (the input is only WordToChar representation) against the same models when we do. We noticed a drop in performance across all models. Notice that all models - even when we do not consider the label – have better performance than the baselines. This also supports the benefits of WordToChar representation.   
\vspace{-2 mm} 

\begin{table}[h!]
\small
\centering
\begin{tabular}{lcc }
Tasks & With Labels & Without Labels \\
\midrule[1pt]
   DIAC+SYN & \textbf{7.70}  & 7.99 \\
    DIAC+POS & \textbf{7.86}  & 7.93 \\
    \midrule[1pt]
    DIAC+SEG+SYN & \textbf{7.70} & 7.93 \\
    DIAC+SEG+POS & \textbf{7.73} & 7.99\\ %change WER and SD
    DIAC+SYN+POS  & \textbf{7.72} & 7.97 \\ %change WER and SD
    \midrule[1pt]
    ALL &  \textbf{7.51}  & 7.91 \\
    \midrule[1pt]
\end{tabular}
  \caption{\label{joint:labelnolabel} WER performance when we do not consider the output labels for the investigated tasks. \textbf{Bold} numbers represent the highest score per row.}
           \vspace{-4mm} 
\end{table}

\if{false}
\begin{table}[h!]
\small
\centering
\begin{tabular}{lcc }
Tasks & With Labels & Without Labels \\
\midrule[1pt]
   DIAC+SYN & \textbf{7.70}  & 7.99 \\
    DIAC+POS & \textbf{7.86}  & 7.93 \\
    \midrule[1pt]
    DIAC+SEG+SYN & \textbf{7.70} & 7.93 \\
    DIAC+SEG+POS & \textbf{7.73} & 7.99\\ %change WER and SD
    DIAC+SYN+POS  & \textbf{7.72} & 7.97 \\ %change WER and SD
    \midrule[1pt]
    ALL &  \textbf{7.63}  & 7.91 \\
    \midrule[1pt]
\end{tabular}
  \caption{\label{joint:labelnolabel} WER performance when we do not consider the output labels for the investigated tasks.}
\end{table}
\fi
	
%\paragraph{CharToWord:}

\subparagraph{Last hidden layer of SEG:} Identifying morpheme boundaries did not increase accuracy as we expected. Therefore, we examined whether information learned from the BiLSTM layer would help us learn morpheme interactions by passing the output of last BiLSTM layer to the diacritic restoration model along with segmentation labels. %Table \ref{segment} shows the different models' performance when we pass information regarding the last BiLSTM layer and when we do not. 
We did not observe any improvements towards predicting accurate diacritics when we pass information regarding the last BiLSTM layer. For ALL, the WER score increased by 0.22\%. Thus, it is sufficient to only utilize the segment labels for diacritic restoration. 

\if{false}
\begin{table}[h!]
\small
\centering
\begin{tabular}{ l|lcccc  }		
   \textbf{Task} & \textbf{With}  & \textbf{WER} &  \textbf{DER} & \textbf{LER/Lex} & \textbf{OOV} \\
    \midrule[1pt]
   \multirow{2}{*}{SEG} & \xmark  & \textbf{8.35} & 2.82 & 5.20/\textbf{5.46} &	33.97    \\
    & \cmark  & 8.43 & \textbf{2.81} & \textbf{5.13}/5.56  & 34.7\\
   \hline
    \multirow{2}{*}{SEG+SYN} & \xmark  & 7.61 & \textbf{2.59} & \textbf{4.65/5.03} & \textbf{31.33}   \\
    & \cmark  & \textbf{7.58} & 2.60 & 4.66/5.06  & 32.30\\
    \hline
      \multirow{2}{*}{SEG+POS} & \xmark  & 7.73 &  \textbf{2.62} & 4.73/\textbf{5.01} & \textbf{31.31}  \\
    & \cmark  & \textbf{7.72} & 2.63	& \textbf{4.72}/5.12 & 31.72 \\
    \hline
    \multirow{2}{*}{ALL} &  \xmark &  \textbf{7.51} & \textbf{2.54} & \textbf{4.54/4.91} & \textbf{31.07} \\
       & \cmark & 7.73 &	2.65 &	4.79/5.08 &	31.87 \\
   \midrule[1pt]
\end{tabular}
  \caption{\label{segment} Performance of different diacritic restoration models when SEG task is considered with (\cmark) and without (\xmark) passing information from the last hidden layer.}
\end{table}

\fi

\subparagraph{Passive and active verbs:} Passivation in Arabic is denoted through diacritics and missing such diacritic can cause ambiguity  in some cases \cite{hermena2015processing,diab2007arabic}. To examine its impact, we further divide verbs in the POS tagset into passive and active, increasing the size  by one. Table \ref{joint:pass} shows the diacritic restoration performance with and without considering passivation. We notice improvements, in some combinations of tasks, across all evaluation metrics compared to the pure POS tagging, showing its importance in diacritic restoration models. 
\vspace{-2 mm} 

\begin{table}[h!]
\small
\centering
\begin{tabular}{l|cc}		
   \textbf{Task} & \textbf{With Pass} & \textbf{Without Pass}  \\
   \midrule[1pt]
    DIAC+POS & \textbf{7.65}  & 7.86  \\
DIAC+SEG+POS & \textbf{7.65}  & 7.73\\
DIAC+SYN+POS & 7.78 & \textbf{7.72}   \\
    ALL & 7.62 &  \textbf{7.51} \\ %%I made up 36.8 just because I cannot currently run the results for that .. 
    \midrule[1pt]
\end{tabular}
  \caption{\label{joint:pass} WER performance for different diacritic restoration models when passivation is considered. \textbf{Bold} numbers represent the highest score per row. %The number of passive verbs in the test set is 425.
  }
\vspace{-4mm} 
\end{table}

\if{false}
\begin{table}[h!]
\small
\centering
\begin{tabular}{l|cc}		
   \textbf{Task} & \textbf{With Pass} & \textbf{Without Pass}  \\
   \midrule[1pt]
    DIAC+POS & \textbf{7.65}/43.3  & 7.86/\textbf{36.0}  \\
DIAC+SEG+POS & \textbf{7.65}/\textbf{35.8}  & 7.73/40.0\\
DIAC+SYN+POS & 7.78/\textbf{36.0} & \textbf{7.72}/38.6   \\
    ALL & 7.62/\textbf{36.2} &  \textbf{7.51}/36.8 \\ %%I made up 36.8 just because I cannot currently run the results for that .. 
    \midrule[1pt]
\end{tabular}
  \caption{\label{joint:pass} WER performance for all words and for passive verbs (separated by /) for different diacritic restoration models when passivation is considered. The number of passive verbs in the test set is 425. \textbf{Bold} numbers represent the highest score per row.}
\vspace{-4mm} 
\end{table}
\fi
%%all verbs (6706) passive verbs(425)
%%43.29411764705882 (DIAC+POS) with passive  36.0 (without passive)   ----- for all verbs with passive 2.7438115120787354 and without passive 2.281538920369818
%%35.76470588235294 with passive 40.0 without passive DIAC+SEG+POS  --- for all verbs with passive 2.2666269012824336 ,, without passive 2.5350432448553533
%%with passive 36.0 without passive 38.588235294117645 --- for all verbs with passive 2.281538920369818 without passive 2.445571130331047 
%%ALL with passive 36.23529411764706 without passive  ?  for all verbs with passive 2.2964509394572024 without passive ? 

%%only characters passed 42.35294117647059 
 
\if{false}
\begin{table*}[h!]
\small
\centering
\begin{tabular}{l|ccccc}		
   \textbf{Task} & \textbf{With Pass} & \textbf{WER} &  \textbf{DER} & \textbf{LER/Lex} & \textbf{OOV}  \\
   \midrule[1pt]
    \multirow{2}{*}{POS} & \xmark  & 7.86 & 2.65 & 4.72/5.20 &	32.28  \\
    & \cmark  & \textbf{7.65} &  \textbf{2.58} &  \textbf{4.65/5.04} &  \textbf{31.16} \\
    \hline
\multirow{2}{*}{SEG+POS} & \xmark  & 7.73 &  2.62 & 4.73/\textbf{5.01} & 31.31  \\
    & \cmark  & \textbf{7.65} & \textbf{2.57} & \textbf{4.71}/5.02 &  \textbf{30.03} \\
    \hline
\multirow{2}{*}{SYN+POS} & \xmark & \textbf{7.72} & \textbf{2.61} & \textbf{4.62}/\textbf{5.06} & 31.05  \\
     & \cmark  & 7.78 & \textbf{2.61} & 4.77/5.07 & \textbf{30.42}   \\
    \hline
    \multirow{2}{*}{ALL} &  \xmark &  \textbf{7.51} & \textbf{2.54} & \textbf{4.54/4.91} & 31.07 \\
       & \cmark  & 7.62 & 2.57 & 4.70/4.99 & \textbf{30.77}\\
   \midrule[1pt]
\end{tabular}
  \caption{\label{joint:pass}Performance of different diacritic restoration models when passivation is considered. \cmark refers to experiments in which we consider passivation as an additional tag while \xmark  refers to experiments in which do not  consider passivation in the tagset.}
\end{table*}
 \fi

 \subparagraph{Level of linguistic information:} 
The joint diacritic restoration model were built empirically and tested against the development set. We noticed that to improve the performance, soft parameter sharing in a hierarchical fashion performs better on diacritic restoration. We experimented with building a joint diacritic restoration model that jointly learns segmentation and diacritics through hard parameter sharing. To learn segmentation with diacritic restoration, we  shared the embedding layer between the two tasks as well as sharing some or all layers of BiLSTM. We got WER on all words (8.53$\sim$9.35) in which no improvements were shown compared to character based diacritic restoration. To learn word based tasks with diacritic restoration, we pass WordToChar representation to the diacritic restoration and/or CharToWord representation for word-based tasks. The best that we could get for both tasks is 8.23\%$\sim$9.6\%; no statistically significant improvements were found. This shows the importance of hierarchical structure for appropriate diacritic assignments.

\paragraph{Qualitative analysis:}
We compared random errors that are correct in DIAC (character-based diacritic restoration) with ALL in which we consider all investigated tasks. Although ALL provides accurate results for more words, it introduces errors in other words that have been correctly diacritized by DIAC. The patterns of such words are not clear. %Table \ref{joint:example} shows some examples in which we have correct diacritic assignments in DIAC and other examples for ALL for the same category of error.
We did not find a particular category that occurs in one model but not the other. Rather, the types and quantity of errors differ in each of these categories. 

\if{false}
\begin{table}[h!]
\small
\centering
\begin{tabular}{lcc}		
\textbf{Category} & \textbf{ALL} & \textbf{DIAC}  \\
   %\hline 
   \midrule[1pt]
   \multirow{4}{*}{Invalid composition} %& \textbf{lAzamahu} & lAzimihu   \\ 
                                        %& \multicolumn{2}{l}{kept jointly with something} \\
                                        %\cline{2-3}
    
                                        & \textbf{waq$\sim$aEatohu} &  waqaEatohu  \\
                                        & \multicolumn{2}{c}{\textit{she made [someone] sign}}\\
                                        \cline{2-3}
   
                                        & nusalamu & \textbf{nusal$\sim$imu}	\\
                                        & \multicolumn{2}{c}{\textit{we greet someone}}\\
                                        %\cline{2-3}

                                         %& $>$usaj$\sim$ilu & 	$>$usoj$\sim$ilu \\
                                        %& \multicolumn{2}{l}{I register}\\ 
                                        
                                        \midrule[1pt]

% \multirow{4}{*}{Passive} & \textbf{Almunotijapi} &	Almunotajapi \\
                                        %& %\multicolumn{2}{c}{\textit{something that is producing}}\\
                                        %\cline{2-3}
   
%                                        & 	$>$uxorajat  & \textbf{$>$axorajat} \\
%                                        & \multicolumn{2}{c}{\textit{[someone] got her out}}\\
                                        
                                        %\midrule[1pt]
                                        
\multirow{4}{*}{Different senses} &  kav$\sim$arat & \textbf{kavurat}\\
                                        & \multicolumn{2}{c}{\textit{she made it greater in quantity}}\\
                                        \cline{2-3}

                                        %& yu\&ominu & \textbf{yu\&am$\sim$inu}\\
                                        %& \multicolumn{2}{l}{he makes [something] secure}\\
                                        %\cline{2-3}
                                        
                                         & \textbf{Almilokiy$\sim$ap}	& Almalakiy$\sim$ap \\
                                        %& \textbf{Almilokiy$\sim$api}	& Almalakiy$\sim$api \\
                                        & \multicolumn{2}{c}{\textit{the possession}}\\
                                        \midrule[1pt]

    \multirow{4}{*}{Case and Mood} %& \textbf{waqaboDihimA}	& waqaboDuhumA \\
                                      %   & \multicolumn{2}{c}{and capturing them ..}\\
                                       % \cline{2-3}
    
                                        & \textbf{maso$>$alapN}	& maso$>$alapF \\
                                         & \multicolumn{2}{c}{\textit{an issue}}\\
                                        \cline{2-3}
    
                                        & funoduqK & \textbf{funoduqi} \\
                                        & \multicolumn{2}{c}{\textit{a hotel}}\\
                                        %\cline{2-3}
    
                                         %& Al$>$umuwru  & Al$>$umuwra \\
                                         %   & \multicolumn{2}{l}{issues}\\
                                         %\hline
                                         \midrule[1pt]
\end{tabular}
  \caption{\label{joint:example} Example of words predicted correctly in ALL but not in DIAC.}
\end{table}

\fi

\if{false}
\subparagraph{Confusion matrix:} The differences between diacritics assigned by DIAC and ALL represented in Figure \ref{joint_confusion}. The two models differ by $\sim$3\% for the diacritics  \textit{N} and \textit{$\sim$K}, $\sim$2\% for the diacritic \textit{$\sim$N}, and $\sim$1\% for the diacritics \textit{u}, \textit{$\sim$}, \textit{K}, and \textit{$\sim$u}. The diacritics \textit{N} and \textit{K} with and without combining them with $\sim$ correspond to the case and mood diacritic marks. The diacritics \textit{u} and \textit{$\sim$u} can be added to the word as a lexical assignment or as a syntactic assignment at the end of the main word. Additionally, \textit{u} and \textit{$\sim$u} usually denote passive verbs for words. The diacritic $\sim$ is equivalent to writing the attached consonants twice and yields a different sense of the word. 
%%SA please add the results of passivization specifically, in all these models, can you measure the passivization diacritic restoration?
%%SA An overarching question: performance on SEG, POS, SYN, report those numbers as well 

\vspace{-4mm} 

\begin{figure}[h!]
%\begin{subfigure}{\textwidth}
  \centering
  \includegraphics[width=\linewidth]{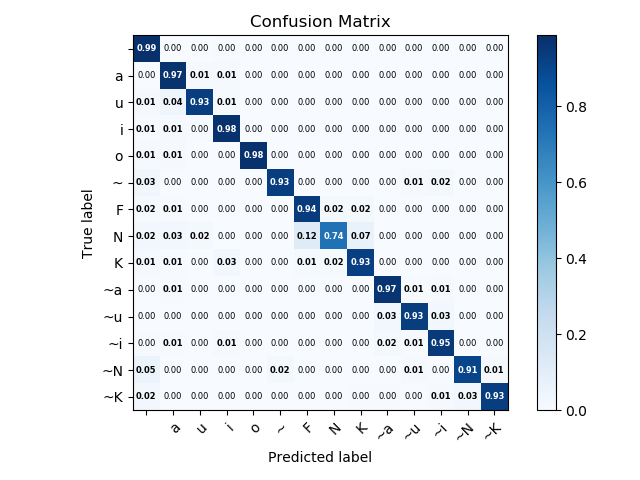}
  %\caption{\label{fig:sub1}DIAC}
%\end{subfigure}%
%\begin{subfigure}{\textwidth}
  %\centering

  \includegraphics[width=\linewidth]{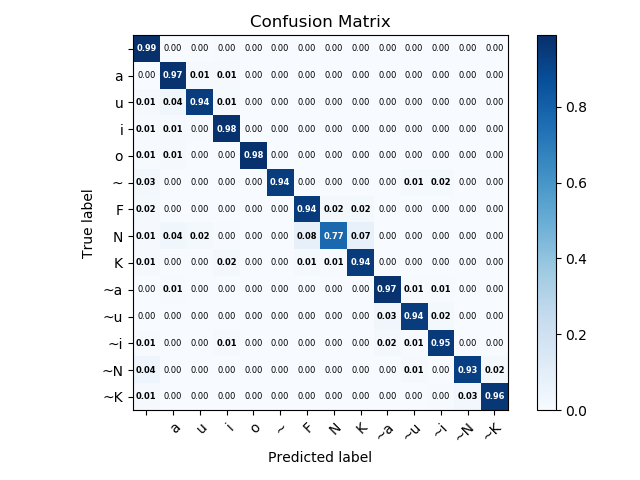}
  %\caption{\label{fig:sub2}All Tasks}
%\end{subfigure}
\caption{\label{joint_confusion}The confusion matrix for DIAC (top) and ALL (bottom). Numbers in the cells represent accuracy.}
\vspace{-4mm} 

\end{figure}
\vspace{-4mm} 

\fi

\paragraph{State-of-the-art Comparison:} 
Table \ref{multi:results} also shows the performance of the state-of-the-art models. %and our  diacritic restoration models. %BiLSTM-CRF \cite{sawsansubword}, which maximize the correct sequence of diacritics, provides comparable performance to our full joint model.  However, BiLSTM-CRF is not efficient for both training and inference. 
ALL model surpass the performance of \newcite{zalmout2017don}.  However, \newcite{zalmout2017don}'s model performs significantly better on OOV words. %This might be due to the way \newcite{zalmout2017don}'s model is formulated (training separate classifiers for extensive set features reflecting morphology in Arabic and using that in addition to the language model output to rank different analysis). 
\newcite{zalmout2019adversarial} provides comparable performance to ALL model.  The difference between their work and that in \cite{zalmout2017don} is the use of a joint model to learn morphological features other than diacritics (or features at the word level), rather than  learning these features individually. %Then they both utilize a language model as well as the output features of their joint model to rank and score all resulted analysis.
\newcite{zalmout2019adversarial} obtained an additional boost in performance  (0.3\% improvement over ours) when they add a dialect variant of Arabic in the learning process, sharing information between both languages. 

\newcite{alqahtani2019investigating} provides comparable performance to ALL and better performance on some task combinations in terms of WER on all and OOV words. The difference between their model and our BASE model is the addition of a CRF (Conditional Random Fields) layer which incorporate dependencies in the output space at the cost of model's computational efficiency (memory and speed). 

\newcite{zalmout2019joint} provides the current state-of-the-art performance in which they build a morphological disambiguation framework in Arabic similar to \cite{zalmout2017don,zalmout2019adversarial}. %which was conducted in parallel with our work. 
They reported their scores based on the development set which was not used for tuning. In the development set, they obtained 93.9\% which significantly outperforms our best model (ALL) by 1.4\%.  Our approach is similar to \cite{zalmout2019joint}. We both follow WordToChar as well as CharToWord input representations discussed in Section \ref{input_representation}, regardless of the specifics. Furthermore, we both consider the morphological outputs as features in our diacritic restoration model. In \newcite{zalmout2019joint}, morphological feature space that are considered is larger, making use of all morphological features in Arabic. Furthermore, \newcite{zalmout2019joint} use sequence-to-sequence modeling rather than sequence classification as ours. Unlike \newcite{zalmout2019joint}, our model is more flexible allowing additional tasks to be added when sufficient resources are available. 

We believe that neither the underlying architecture nor the consideration of all possible  features were the \textit{crucial} factor that led to the significant reduction in WER performance. Rather, morphological analyzers is crucial  in such significant improvement. As a matter of fact, in \newcite{zalmout2019joint}, the performance significantly drops to 7.2 when they, similar to our approach, take the highest probabilistic value as a solution. Thus, we believe that the use of morphological analyzers enforces valid word composition in the language and filter out invalid words (a side effect of using characters as input representation). This also justifies the significant improvement on OOV words obtained by \cite{zalmout2017don}.  Thus, we believe that a global knowledge of words and internal constraints within words are captured.

\if{false}
\begin{table*}[h!]
\small
\centering
\begin{tabular}{ l|l|l  }		
   \textbf{Task} & \textbf{Model}  & \textbf{Accuracy}\\
  \midrule[1pt]
   \multirow{4}{*}{Diacritic Restoration}  &  \newcite{zalmout2017don} & 91.70 \\
                                           %& BiLSTM+45 \cite{} & 91.80 \\ 
                                           %& TCN \cite{} & 89.80\\
                                           & BiLSTM\_CRF & 92.40 \\
                                           & BiLSTM+300 (our base) & 91.50 \\ 
                                           & ALL & \textbf{92.49}\\
                                           &ALL with passive & 92.38\\
                                           & \cite{zalmout2019adversarial} & 92.5\%\\
     \midrule[1pt]
        \multirow{2}{*}{Word Segmentation}  &  \newcite{zalmout2017don} & 99.60\\
                                            &    BiLSTM (base)        & 99.88 \\
    \midrule[1pt]
   \multirow{1}{*}{POS Tagging} & BiLSTM (base) & 97.15 \\ 

   \midrule[1pt]

   \multirow{2}{*}{Syntactic Diacritization} & \newcite{hifny2018hybrid} & 94.70 \\
                                            & BiLSTM (base) & 94.22 \\
   \midrule[1pt]
\end{tabular}
  \caption{\label{baselines} The performance of our models with the previous state-of-the-art models.}
\end{table*}
\fi

%In both \cite{zalmout2019joint,zalmout2019adversarial}, diacritic restoration is learned  within morphological analysis framework as in the case of \cite{zalmout2017don}. 

%\if{false}
\subparagraph{Auxiliary tasks:}
We compared the base model of the auxiliary tasks to the state-of-the-art (SOTA). For SEG,  BiLSTM model has comparable performance to that in   \cite{zalmout2017don} (SEG yields 99.88\% F1 compared to SOTA 99.6\%). For POS, we use a shallower tag set (16 number of tags compared to $\sim$70) than typically used in previous models hence we do not have a valid comparison set. For SYN, we compare our results with \cite{hifny2018hybrid} which uses a hybrid network of BiLSTM and Maximum Entropy to solve syntactic diacritization. The SYN yields results comparable to SOTA (our model performs 94.22 vs. SOTA 94.70). 
                                           % & BiLSTM  & 94.22SEG and SYN. 

%\vspace{-2mm}
%\begin{table}[h!]
%\small
%\centering
%\begin{tabular}{ l|l|l  }		
 %  \textbf{Task} & \textbf{Model}  & \textbf{Accuracy}\\
 % \midrule[1pt]
  %      \multirow{2}{*}{SEG}  &  \newcite{zalmout2017don} & 99.60\\
   %                                         &    BiLSTM         & 99.88 \\
%    \midrule[1pt]
 %  \multirow{1}{*}{POS} & BiLSTM & 97.15 \\ 

 %  \midrule[1pt]

  % \multirow{2}{*}{SYN} & \newcite{hifny2018hybrid} & 94.70 \\
   %                                         & BiLSTM  & 94.22 \\
   %\midrule[1pt]
%\end{tabular}
%  \caption{\label{baselines} The performance of our models with the previous state-of-the-art models.}
%  \vspace{-4mm}
%\end{table}
%\fi

%%SA move the related work after the discussion and compare and contrast them with your models  DONE

\section{Related Work}
The problem of diacritization has been addressed using classical machine learning approaches (e.g. Maximum Entropy and Support Vector Machine) \cite{zitouni2009arabic,pasha2014madamira} or neural based  approaches for different languages that include diacritics such as Arabic, Vietnamese, and Yoruba.  Neural based approaches yield state-of-the-art performance for diacritic restoration by using Bidirectional LSTM or temporal convolutional networks \cite{zalmout2017don,orife2018attentive,sawsantcn, alqahtani2019investigating}.

%%SA introduce the different types of arabic diacritization lexicalized which are pattern related (templatic) while syntactic diacritics that tend to be word final and reflect syntactic case -- provide examples.  DONE
Arabic syntactic diacritization has been consistently reported to be difficult, degrading the performance of full diacritic restoration \cite{zitouni2006maximum,habash2007determining,said2013hybrid,shaalan2009hybrid,shahrour2015improving,darwish2017arabic}. To improve the performance of syntactic diacritization or full diacritic restoration in general, previous studies followed different approaches. Some studies  separate lexical from syntactic diacritization \cite{shaalan2009hybrid,darwish2017arabic}. Other studies consider additional linguistic features such as POS tags and word segmentation (i.e. tokens or morphemes) \cite{ananthakrishnan2005automatic,zitouni2006maximum,zitouni2009arabic,shaalan2009hybrid}. 

\citet{hifny2018hybrid} addresses syntactic diacritization by building BiLSTM model in which its input embeddings are augmented with manually generated features of context, POS tags, and word segments. \citet{rashwan2015deep} use deep belief network to build a diacritization model for Arabic that focuses on improving syntactic diacritization and build sub-classifiers based on the analysis of a confusion matrix and POS tags.   

Regarding incorporating linguistic features into the model, previous studies have either used morphological features as a preprocessing step or as a ranking step for building diacritic restoration models.  As a preprocessing step, the words are converted to their constituents (e.g. morphemes, lemmas, or $n$-grams)  and then diacritic restoration models are built on top of that \cite{ananthakrishnan2005automatic,sawsansubword}. \citet{ananthakrishnan2005automatic} use POS tags to improve diacritic restoration at the syntax level assuming that POS tags are known at  inference time.

As a ranking procedure, all possible analyses of words are generated and then the most probable analysis is chosen \cite{pasha2014madamira,zalmout2017don,zalmout2019adversarial,zalmout2019joint}. \citet{zalmout2017don} develop a morphological disambiguation model to determine Arabic morphological features including diacritization. They train the model using BiLSTM and consult with a LSTM-based language model as well as other morphological features to rank and score the output analysis. Similar methodology can be found in \cite{pasha2014madamira} but  using Support Vector Machines. This methodology shows better performance on out of vocabulary (OOV) words compared to pure character models.

\section{Discussion \& Conclusion}
We present a diacritic restoration joint model that considers the output distributions for different related tasks to improve the performance of diacritic restoration. Our results shows statistically significant improvements across all evaluation metrics. %In addition, it shows improvements in certain types of  diacritics. 
This shows the importance of considering additional linguistic information at  morphological and/or sentence levels. %Not only syntactic and morphological information are considered, 
Including semantic information through pretrained word embeddings within the diacritic restoration model also helped boosting the diacritic restoration performance. Although we apply our joint model on Arabic, this model provides a framework for other languages that include diacritics whenever resources become available. Although we observed improvements in terms of generalizing beyond observed data when using the proposed linguistic features, the OOV performance is still an issue for diacritic restoration. 

\bibliography{acl2020}
\bibliographystyle{acl_natbib}

\end{document}